\documentclass{article}
\usepackage{graphicx}
\usepackage{float}
\usepackage{enumitem}
\usepackage{subcaption}
\usepackage{subcaption}
\usepackage{todonotes}
\usepackage{booktabs}
\usepackage{amsmath}
\usepackage{tabularx}
\usepackage{enumitem}
\usepackage{amssymb} 

\usepackage[preprint]{neurips_2026}

\usepackage[utf8]{inputenc} 
\usepackage[T1]{fontenc}    
\usepackage{hyperref}       
\usepackage{url}            
\usepackage{booktabs}       
\usepackage{amsfonts}       
\usepackage{nicefrac}       
\usepackage{microtype}      
\usepackage{xcolor}         

\title{Are You Thinking What I’m Thinking? : Examining Conceptual Separation in Neural Architectures}
\makeatletter
\renewcommand{\@noticestring}{}
\makeatother

\author{
Jaee Ponde\thanks{Equal contribution. The work was done at Ashoka University.} \\
\textit{Truth Audit Labs} \\
\texttt{jaee@truthauditlabs.ai}
\And
Roshni Agarwal\footnotemark[1] \\
\textit{Karya AI} \\
\texttt{roshni.agarwal@karya.in}
\And
Subhashis Banerjee\thanks{Department of Computer Science and the Centre of Digitalisation, AI, and Society, Ashoka University, Sonepat, Haryana, India}\\
\textit{Ashoka University} \\
\texttt{suban@ashoka.edu.in}
}

\begin{document}

\raggedbottom
\setlength{\textfloatsep}{8pt plus 2pt minus 2pt}

\maketitle

\begin{abstract}
Neural networks are increasingly employed to identify both well-defined and ambiguous concepts, yet output-level metrics reveal little about how those concepts are represented internally. Our study asks if these networks exhibit \textit{conceptual separation}: if examples of the same concept form coherent representations, and whether related concepts lie closer together in the representation space. We examine this conceptual organisation in Convolutional Neural Networks (CNNs) and Large Language Models (LLMs) through geometric and distributional analysis of their internal activations. In CNNs, familiar ImageNet concepts form coherent and semantically ordered representations, while this coherence weakens for unseen concepts and suffers within-class domain shift. In LLMs, clearly distinct domains remain well separated, related subdomains move closer together, and the distinction between ambiguous topics collapses at both the mean and covariance level. These results suggest that conceptual separation can reveal structure that output accuracy alone cannot, and may serve as a useful diagnostic of how robustly a model represents the concepts it is asked to identify. Code and data available on \href{https://github.com/JaeeRoshniCapstoneProject/Are-You-Thinking-What-I-m-Thinking-Examining-Conceptual-Separation-in-Neural-Architectures}{GitHub}.
\end{abstract}

\section{Introduction}

Humans have an innate understanding of conceptual separation. We recognise that a cat and a dog are different classes, while also understanding that they are more similar to each other than either is to a car. This notion of proximity becomes complicated as concepts become less well defined. Subtopics within the same domain may be difficult to separate cleanly, and subjective concepts such as hate speech often have no universally agreed boundary; even human annotators can disagree substantially on the same examples \cite{davani2022dealing,fleisig2023majority}. Conceptual separation, therefore, is not simply about whether two labels are different, but also about how clearly and consistently that distinction can be made.

Neural networks are increasingly deployed to classify both well-defined and ambiguous concepts, despite us knowing relatively little about whether they preserve such conceptual structure. They are evaluated largely through output-level metrics that can coexist with shortcut learning, underspecification, and poor behaviour under distribution shift \cite{geirhos2020shortcut,damour2022underspecification}. This motivates us to look beyond benchmark performance and test a model's understanding of the underlying task. A natural place to do so is the activation space, which has been widely used to study what neural networks encode, how representations evolve across layers, and how similar representations are across models \cite{raghu2017svcca,kornblith2019similarity}.  We study whether the geometry of this activation space reveals the relative proximity and separation of concepts themselves.

We present a set of empirical experiments examining conceptual separation in CNNs and LLMs. CNNs are a natural starting point because visual classification benchmarks
such as ImageNet provide explicit class labels and a controlled setting
for studying representational separation \cite{russakovsky2015imagenet}. This gives us a controlled setting in which to first ask whether familiar classes are separated as expected. From there, we extend the same question to language, where concept boundaries are often less clear. We study distinct domains, related technical subdomains, and finally hate versus non-hate speech, which are more ambiguous. Across both modalities, we use multiple geometric and distributional views of the activation space.

Our results suggest that internal activations provide a view beyond output labels, revealing how well a concept is represented relative to others inside the model. This structure can degrade across unseen, shifted, or ambiguous concepts, motivating the need for closer inspection of internal representations before real-world deployment. Our contributions are:
\begin{itemize}
\item We introduce an architecture-aware empirical framework for studying conceptual separation directly in neural activation spaces.
\item In CNNs, we show that familiar classes exhibit coherent internal structure, which weakens not only for unseen concepts but also domain shifts within a class.
\item In LLMs, we study a progression from clearly distinct domains to ambiguous domains, finding a corresponding degradation in representational separation.
\item We extend the LLM analysis beyond mean-level geometry to covariance structure, testing whether distinctions that disappear in the first moment remain visible in the second.
\end{itemize}

\section{Related Work}

Our work is related to three strands of literature. First, \textit{concept-based interpretability} seeks to explain neural networks in terms of human-meaningful concepts rather than individual features. TCAV introduced Concept Activation Vectors to quantify the sensitivity of model predictions to user-defined concepts \cite{kim2018tcav}, while Concept Bottleneck Models explicitly route predictions through human-interpretable concept variables \cite{koh2020concept}. Our goal is complementary: rather than asking whether a particular concept influences a prediction, we study how multiple concepts are organised relative to one another in the model's activation space.

Second, a large body of work studies the structure and similarity of learned representations. Representational Similarity Analysis compares pairwise dissimilarity structure across representational spaces \cite{kriegeskorte2008representational}, while methods such as SVCCA and CKA quantify similarity between neural representations across layers or models \cite{raghu2017svcca,kornblith2019similarity}. In NLP, probing classifiers are widely used to test whether linguistic or semantic properties can be decoded from internal representations \cite{belinkov2019analysis,belinkov2022probing}. We instead focus on \textit{conceptual separation}: whether concepts form coherent internal representations and how close different concepts are to one another.

Finally, our diagnostic framing connects to work on out-of-distribution detection and domain generalisation, which studies how model behaviour changes under inputs that differ from the training distribution \cite{hendrycks2017baseline,hendrycks2021many,zhou2023domain}. Rather than relying only on output confidence or prediction accuracy, we examine shifts in activation distributions as evidence of whether conceptual structure remains stable under unseen or shifted inputs.

\section{Experiments on Convolutional Neural Networks}

We first study conceptual separation in image classifiers, where class boundaries are relatively clear. We ask whether images from the same class are represented coherently and whether some concepts are ``closer'' to one another than others.

\subsection{Setup}

We study two ImageNet-trained CNN architectures: ResNet-50 \cite{he2016deep} and MobileNetV2 \cite{sandler2018mobilenetv2}. For each input image $x$, we extract the activation vector from the layer immediately preceding the final classifier,
$z(x) \in \mathbb{R}^{d}$,
with $d=2048$ for ResNet-50 and $d=1280$ for MobileNetV2 as deeper CNN layers tend to encode richer features \cite{yosinski2014transferable}. We treat this vector as the model's representation of the image and study whether images belonging to the same concept occupy more similar regions of this representation space than images belonging to different concepts.

Our initial experiments use three familiar ImageNet concepts--cats, dogs, and cars. We subsequently consider unseen concepts such as rangoli and microscopy, geographic variation in road images from India and Turkey, and pose variation within the cat class. Images are drawn from publicly available Kaggle datasets \cite{magesh_rangoli,jacob_catdog,chakrawarty_idd,
saridogan_traffic,yildirim_microscopy_kaggle,prondeau_carconnection}. We use 100 randomly sampled images of each class across experiments, examples available in Appendix \ref{app:dataset_examples}.

We explore several metrics to check conceptual separation. PCA is applied to the raw activation vectors after mean-centering, cosine distance normalises by vector magnitude, and Mahalanobis distance accounts for the within-class covariance structure. Finally, we use KL divergence \cite{kullback1951information} to compare the
distribution of activation across neurons. This is inspired by prior
work on KL-based measures of test-sample fit \cite{meegahapola2019pad}. We adapt this idea to study conceptual
separation: both the coherence of examples within a concept and the relative
proximity between different concepts. For each concept $c$, we construct an activation blueprint by averaging the representations of $N=100$ reference images, $\bar{z}_c=\frac{1}{N}\sum_{i=1}^{N} z(x_i)$, and normalising it to sum to one, $Q_c(k)=\bar{z}_c(k)/\sum_j \bar{z}_c(j)$. A test image is normalised in the same way, $P_x(k)=z(x)_k/\sum_j z(x)_j$, and compared with the concept blueprint using KL divergence:

\begin{equation}
D_{\mathrm{KL}}(P_x\|Q_c)
=
\sum_k P_x(k)
\log\left(
\frac{P_x(k)}
{Q_c(k)+\epsilon}
\right),
\qquad \epsilon=10^{-8}
\end{equation}

Here, $\epsilon$ is added for numerical stability.
Intuitively, $Q_c$ represents the typical relative pattern of neuron activation for concept $c$. A low KL divergence indicates that a test image produces a similar activation pattern, while a high value indicates that its representation departs from the concept blueprint. For inter-class comparisons, we compute KL divergence in both directions and report their average, i.e., $\frac{1}{2}\left[
D_{\mathrm{KL}}(Q_c\|Q_d)
+
D_{\mathrm{KL}}(Q_d\|Q_c)
\right]$, to obtain a symmetric measure of separation between concepts. This allows us to study both separation between concepts and coherence within a concept under unseen or shifted inputs.
\subsection{In-Distribution Concept Separation}

Our first experiment is concerned with the preservation of simple semantic structure between familiar ImageNet classes: cats, dogs, and cars. 

As a first order analysis, we use these activations directly for Euclidean distance and PCA, cosine distance and Mahalanobis distance. Across both ResNet-50 and MobileNetV2, intra-class distances are smaller than inter-class distances. Cats and dogs are also closer to each other than either is to cars, indicating that the activation space preserves the semantic structure we would expect. The corresponding numerical results and methodology are reported in Appendix~\ref{app:cnn_distances}.

\begin{figure*}[t]
    \centering
    \includegraphics[width=0.85\textwidth]{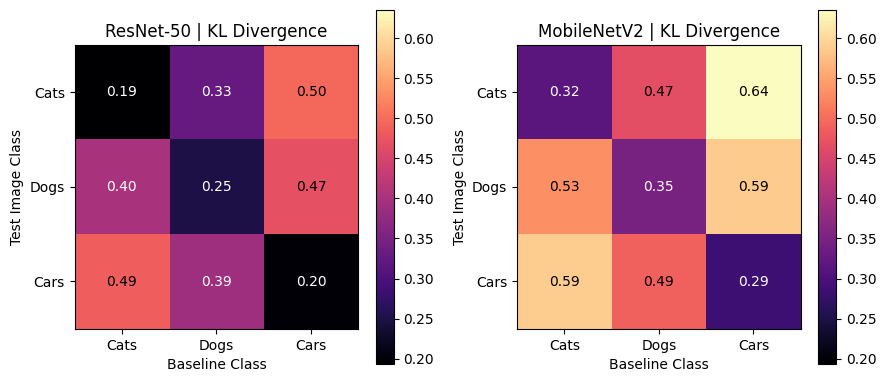}
    \caption{
    KL divergence between test-image activation distributions and class activation blueprints for ResNet-50 (left) and MobileNetV2 (right).
    Lower diagonal values show that images are closest to their own class blueprint.
    }
    \label{fig:kl_heatmap}
\end{figure*}

\begin{figure*}[t]
    \centering
    \includegraphics[width=\textwidth]{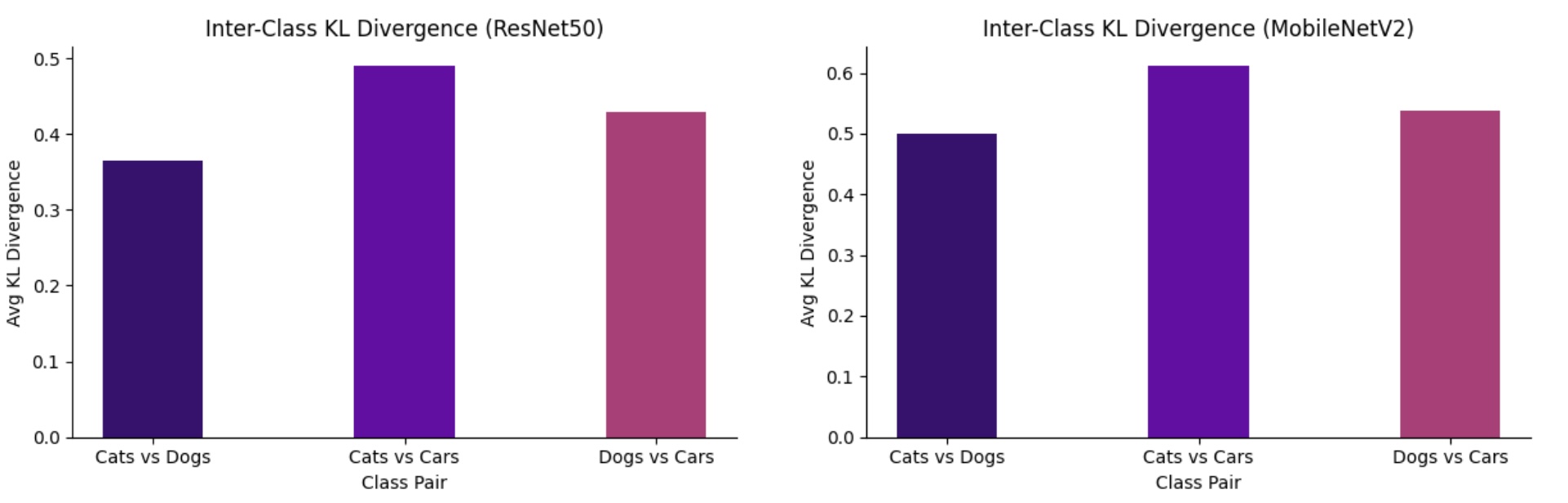}
    \caption{
    Mean inter-class KL divergence for ResNet-50 (left) and MobileNetV2 (right).
    Both models recover the same semantic ordering: cats and dogs have the most
    similar activation patterns, while cats and cars are the most distinct.
    }
    \label{fig:kl_interclass}
\end{figure*}

For both architectures, KL divergence is lowest when an image is compared with its own class blueprint (Figure~\ref{fig:kl_heatmap}). The inter-class comparison also preserves the expected semantic ordering: cats and dogs have more similar activation distributions than cats and cars (Figure~\ref{fig:kl_interclass}). Unlike point-wise distances, KL divergence captures the full pattern of relative neuron activation, giving us a more holistic measure of representational similarity. This provides evidence that the models are not only identifying concepts as distinct classes, but are also organising them relationally, with some concepts represented as more alike than others.

\subsection{Out-of-Distribution Experiments}

Out-of-distribution settings admit several definitions and evaluation protocols \cite{yang2024generalized}; here, we use the term in a strictly semantic sense, referring to visually coherent concepts that are not represented as explicit ImageNet classes. Detecting such unfamiliar inputs is important for reliable deployment, since models may encounter inputs that differ from those seen during training \cite{hendrycks2017baseline,yang2024generalized}.

We test two human-distinguishable concepts that are not explicit ImageNet classes: Rangoli, a South Asian decorative art form made from geometric and floral patterns, and microscopy images. Rangoli and microscopy therefore serve as semantically OOD concepts: while the models may have encountered visually similar features during training, they were not trained to recognise these concepts as such. We repeat the same activation-blueprint experiment to ask whether unfamiliar concepts still produce coherent internal
representations.

\begin{figure*}[t]
    \centering
    \includegraphics[width=\textwidth]{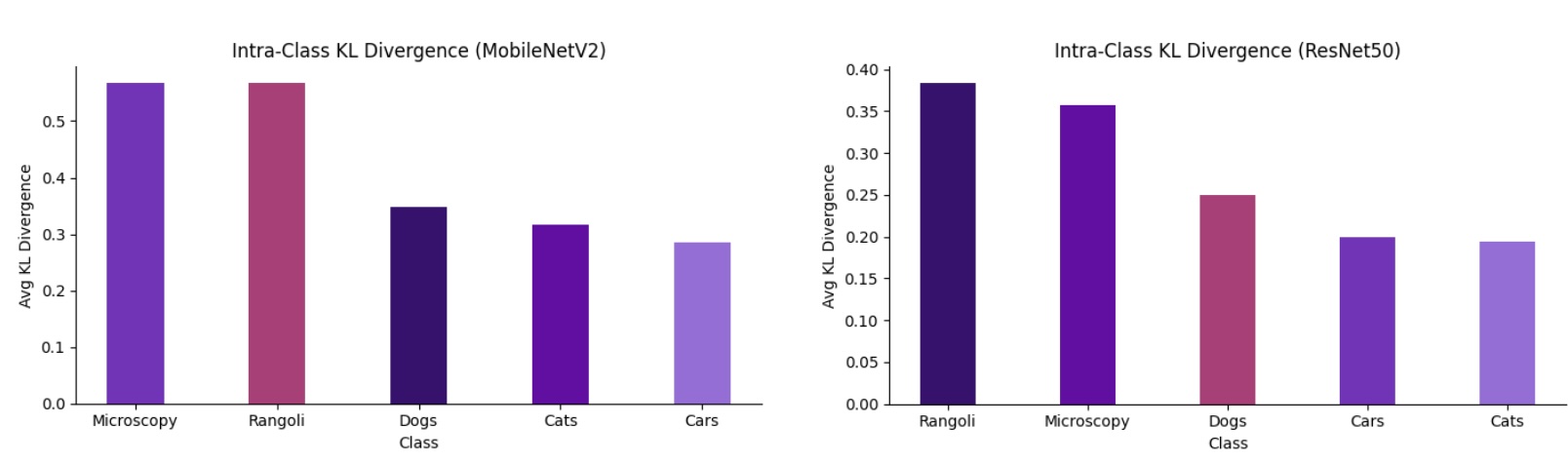}
    \caption{
    Intra-class KL divergence for ImageNet and out-of-distribution concepts.
    OOD images produce substantially less consistent activation
    patterns than Imagenet Classes. MobileNetV2 (left) and ResNet-50 (right).
    }
    \label{fig:ood_kl}
\end{figure*}

Both concepts show substantially higher intra-class KL divergence than the familiar ImageNet classes (Figure~\ref{fig:ood_kl}). This suggests that different images from the same human-defined concept do not produce a single, stable activation pattern in the way cats, dogs, and cars do. This does not imply that the model fails to extract useful visual features; rather, those features appear less consistently organised at the concept level. We therefore treat intra-class KL divergence here as a lightweight diagnostic of representational coherence under unfamiliar inputs.

We also examine the models' ImageNet predictions for these images. Unlike the in-distribution classes, Rangoli and microscopy images are mapped across many unrelated ImageNet labels, providing a behavioural counterpart to the activation-space result. Results and further discussion are provided in Appendix~\ref{app:ood_predictions}.

\subsection{Within-Class Distribution Shifts}

We next ask whether KL divergence can detect differences within a single concept. This is important because models deployed in the real world often encounter domain shifts, where the underlying class remains unchanged but visual conditions differ, and such shifts can substantially affect model robustness \cite{hendrycks2021many,zhou2023domain}. In these experiments, we use 200 images for the base class to ensure distinct blueprint and held-out test sets. 

\begin{figure}[H]
    \centering
    \includegraphics[width=1\columnwidth]{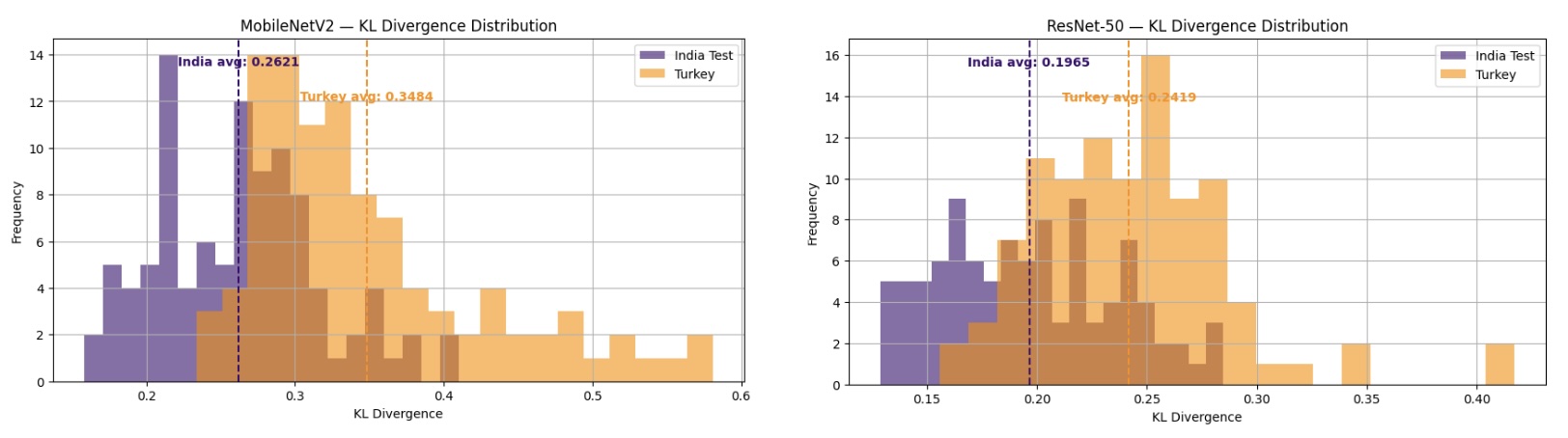}
    \caption{
    KL divergence of Indian and Turkish road images from an Indian-road activation blueprint.
    Turkish road images show a clear drift from the Indian baseline.
    }
    \label{fig:road_shift}
\end{figure}

We compare street images from India and Turkey. Although both sets depict roads, they differ in visual context, including vegetation, vehicles, road markings, and surrounding infrastructure. Using an Indian-road activation blueprint, Turkish road images show consistently higher KL divergence than held-out Indian images (Figure \ref{fig:road_shift}) suggesting that the metric is sensitive to within-class distribution shift.  In this sense, the blueprint comparison can serve as a simple pre-deployment check: given a model trained on a particular class distribution, we can test whether new examples of that same class induce substantially different internal representations before relying on the model in a new environment.

\begin{figure}[H]
    \centering    \includegraphics[width =1\columnwidth]{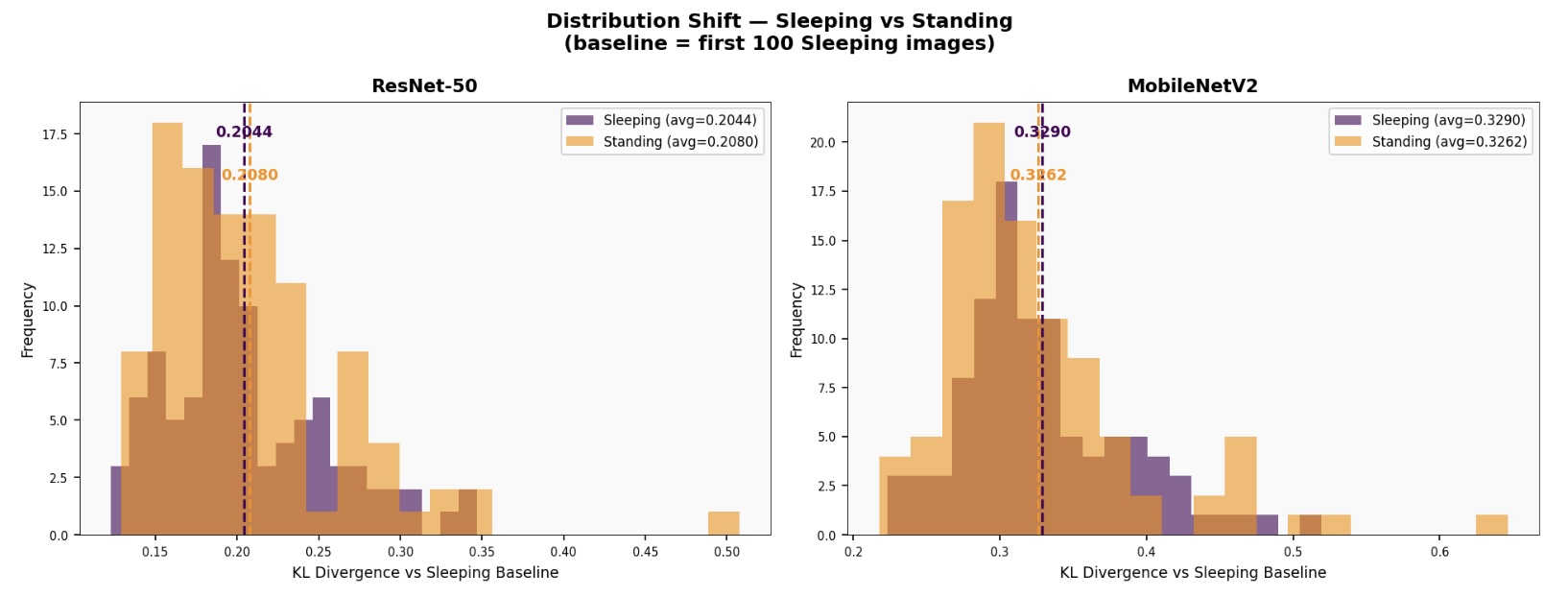}
    \caption{
    KL divergence of sleeping and standing cat images from a sleeping-cat activation blueprint.
    The two distributions largely overlap in both models.
    }
    \label{fig:cat_pose}
\end{figure}

We then test a different attribute within the cat class: sleeping versus standing cats. Using a sleeping-cat activation blueprint, the KL divergence distributions for sleeping and standing cats are nearly identical (Figure~\ref{fig:cat_pose}). In contrast to the road experiment, changing pose does not produce a clear shift in the activation distribution. The models therefore appear to treat sleeping and standing cats as similarly coherent instances of the same concept, at least under this metric. 

The two experiments suggest that not all within-class variation is reflected equally in representation space: geographic and contextual changes in road scenes are detectable, while pose differences within cats are largely absorbed into the broader class representation. One possible explanation is that pose is not a distinction the ImageNet objective explicitly encourages the models to separate, whereas road scenes differ along several visual dimensions that may already influence features learned during ImageNet training.

\section{Experiments on Large Language Models}

The CNN experiments established that familiar concepts produce coherent internal representations while unfamiliar or shifted concepts do not. We now ask whether the same principle holds in language, where concept boundaries are often less perceptual and more socially constructed. We study three concept pairs that form a deliberate progression of human-separability:

\begin{enumerate}[nosep]
    \item \textbf{Computer Science vs.\ Shakespeare}--maximally distinct domains that no literate reader could confuse, serving as a positive control.
    \item \textbf{Information Security vs.\ Theory of Computation}--sub-domains within computer science that a specialist can distinguish but that share technical vocabulary and register.
    \item \textbf{Hate speech vs.\ no-hate speech}--a socially contested boundary where even trained annotators systematically disagree on class membership \cite{davani2022dealing,fleisig2023majority}.
\end{enumerate}

This progression tests whether the degradation of representational coherence observed for visual concepts extends to textual concepts, and whether it tracks the degree of human-perceived conceptual ambiguity.

\subsection{Setup}

We extract sentence embeddings from two architecturally contrasting models:
GPT-OSS-20B ($N{=}2880$) \cite{openai2025gptoss}, an open-weight autoregressive transformer,
and BERT-large-uncased \cite{devlin2019bert}, a 340M-parameter bidirectional masked-language encoder ($N{=}1024$). GPT's causal attention produces representations conditioned only on prior context, while BERT's bidirectional attention integrates the full sentence simultaneously. 

For each model, we mean-pool the final-layer hidden states (weighted by the attention mask) to produce a single sentence embedding $\mathbf{f}(x) \in \mathbb{R}^{N}$. For the academic pairs, 100 sentences per class are drawn from Wikipedia and academic corpora. For hate speech, 100 sentences per class come from a single corpus of tweets from the 2020 US presidential election~\cite{grimminger2021hate}, ensuring that any observed difference reflects the concept itself rather than platform or stylistic confounds. Dataset details are in Appendix~\ref{app:hate_data}.

We report pairwise Euclidean and cosine \emph{separability ratios} (inter-class distance divided by mean intra-class distance; a value of 1.0 indicates no separation) and 3D PCA projections (Figure~\ref{fig:pca3d_llm}). Full distance matrices and auxiliary analyses are provided in Appendix~\ref{app:llm_distance}.

\subsection{First-Moment Analysis}
\begin{table}[H]
\centering
\caption{Summary of representational separability across three concept pairs and two models. Cosine separability ratio measures first-moment (centroid-level) separation; normalised Frobenius norm $\tilde{F}$ and its permutation-test $p$-value measure second-moment (covariance-level) separation. Values are reported as GPT-OSS-20B / BERT-large-uncased.}
\label{tab:llm_summary}
\small
\begin{tabular}{lcccc}
\toprule
\textbf{Concept Pair} & \textbf{Cosine Ratio} & $\tilde{F}$ & $p$\textbf{-value} \\
\midrule
CS vs.\ Shakespeare       & 1.406 / 1.487  & 1.238 / 1.267 & $<$0.001 / $<$0.001 \\
InfoSec vs.\ ToC           & 1.358 / 1.505 & 1.268 / 1.195 & $<$0.001 / $<$0.001 \\
Hate vs.\ No-Hate          & 1.012 / 1.012  & 0.416 / 0.492 & 0.025 / 0.130 \\
\bottomrule

\end{tabular}
\end{table}

Table~\ref{tab:llm_summary} summarises the first-moment and second-moment results across all three pairs and both models. As the conceptual distinction becomes more ambiguous from a human perspective, the geometric metrics reflect the same loss of separability.

\begin{figure}[!t]
    \centering
    \includegraphics[width=0.8\linewidth]{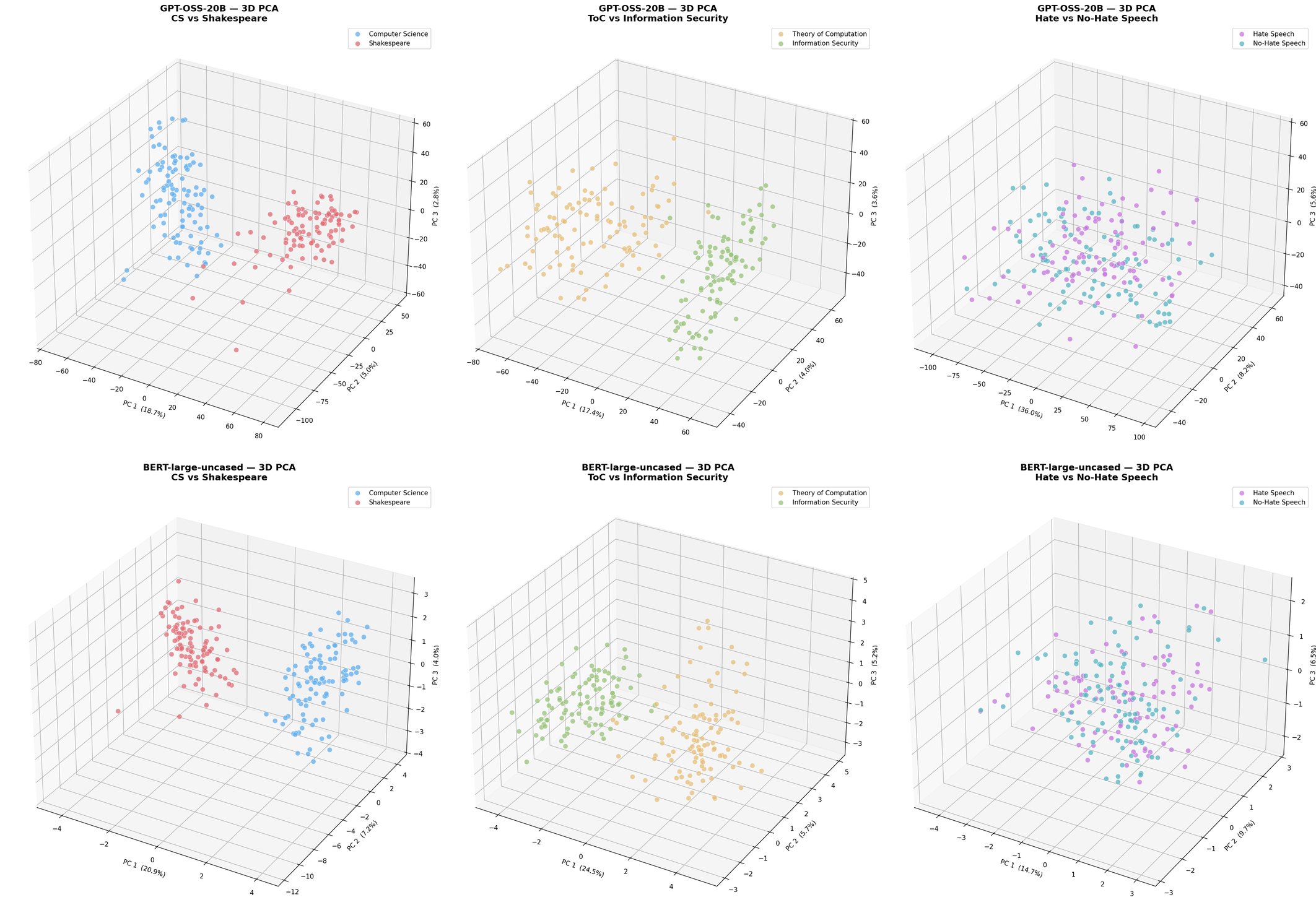}
    \caption{3D PCA projections of sentence embeddings. Top row: GPT-OSS-20B; Bottom row: BERT-large-uncased. Columns (left to right): CS vs.\ Shakespeare (well-separated), InfoSec vs.\ ToC (moderate overlap), and Hate vs.\ No-Hate Speech (complete geometric entanglement).}
    \label{fig:pca3d_llm}
\end{figure}

\paragraph{Academic pairs.}
For CS vs.\ Shakespeare, both models show a clear separation, 3D PCA projections show two distinct spatial clusters in the leading components (Figure~\ref{fig:pca3d_llm}, top row) and the cosine separability ratio exceeds 1.4 (Table \ref{tab:llm_summary}).
For InfoSec vs.\ Theory of Computation, separation remains significant. Cosine ratios drop slightly (Table \ref{tab:llm_summary}) but still evident of distinct clusters, consistent with the hypothesis that conceptually closer domains produce less geometrically separated representations. 

\paragraph{Hate vs.\ no-hate speech.}
For hate speech vs.\ no-hate speech, first-moment separation collapses entirely. The cosine separability ratio falls to effectively unity in both models. 3D PCA projections (Figure~\ref{fig:pca3d_llm}, bottom row) display complete spatial entanglement. 

To quantify distributional differences beyond centroids, we compute a per-neuron KL divergence between classes. Because LLM hidden states can take negative values, we cannot treat an embedding vector as a probability distribution. Instead, for each neuron $j$, we bin its activation values across all $M$ sentences in each class into a 50-bin histogram with shared global bin edges, yielding a probability mass function $\tilde{H}_{c,j}$. The per-neuron KL divergence is:
\begin{equation}
    \mathrm{KL}(\tilde{H}_{c,j} \| \tilde{H}_{d,j}) = \sum_{b} \tilde{H}_{c,j}(b) \cdot \log \frac{\tilde{H}_{c,j}(b)}{\tilde{H}_{d,j}(b)}
\end{equation}
Averaging across all $N$ neurons yields a single scalar per class pair. This captures each neuron's full distributional shape without parametric assumptions, though it discards inter-neuron correlations (which the covariance analysis in Section \ref{second} addresses).

A six-category KL divergence analysis reveals that the per-neuron KL divergence between hate speech and no-hate speech is the smallest off-diagonal value in the entire $6 \times 6$ matrix--smaller even than between Information Security and Theory of Computation. From the models' representational perspective, hate and no-hate speech are more similar than any two academic categories, including those from the same technical discipline (Appendix~\ref{app:llm_distance}).

\subsection{Second-Moment Analysis}
\label{second}

An absence of first-moment separability does not preclude all representational structure: the class centroids may coincide while the \emph{shapes} of their activation distributions differ. A neuron whose activations are consistently high for hate speech but highly variable for no-hate speech would produce identical centroids but different covariance matrices. We therefore test whether the classes differ at the second moment.

\begin{figure}[H]
    \centering
    \includegraphics[width=\linewidth]{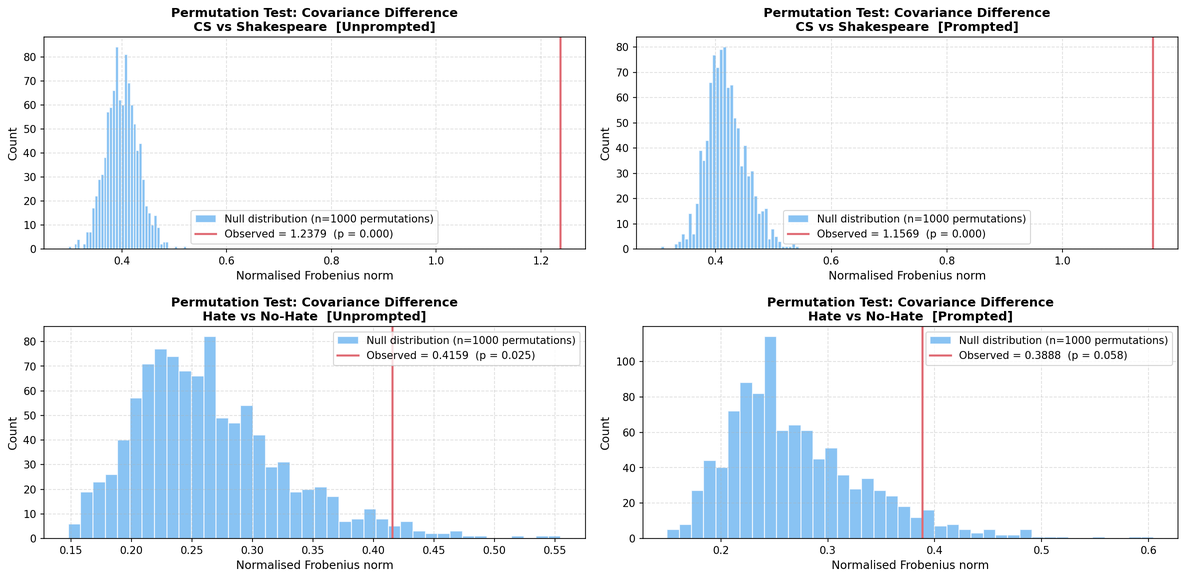}
    \caption{Permutation test null distributions (GPT-OSS-20B, 1000 random relabellings) of the normalised Frobenius norm $\tilde{F}$. Top row: CS vs.\ Shakespeare; bottom row: Hate vs.\ No-Hate Speech. Left: unprompted; right: prompted. The red vertical line marks the observed $\tilde{F}$.}
    \label{fig:frobenius}
\end{figure}

We model each class's distribution in PCA-50 space as a multivariate Gaussian $\mathcal{N}(\boldsymbol{\mu}, \boldsymbol{\Sigma})$, estimated from 100 sentences per class. The normalised Frobenius norm,
\begin{equation}
    \tilde{F} \;=\; \frac{\lVert \boldsymbol{\Sigma}_1 - \boldsymbol{\Sigma}_2 \rVert_F}{\tfrac{1}{2}\bigl(\lVert \boldsymbol{\Sigma}_1 \rVert_F + \lVert \boldsymbol{\Sigma}_2 \rVert_F\bigr)}\,,
    \label{eq:frobenius}
\end{equation}
measures how different the two classes' covariance matrices are, scaled by their average magnitude. A value of $\tilde{F} = 0$ means the two classes spread their activations in exactly the same way; A value of $\tilde{F}=0$ indicates identical covariance matrices, while larger
values indicate greater covariance dissimilarity relative to their overall magnitude.
We assess whether the observed difference is meaningful using a permutation test. We randomly shuffle the class labels 1000 times, recompute $\tilde{F}$ each time, and build a null distribution of values expected by chance. If the observed $\tilde{F}$ (red line in Figure~\ref{fig:frobenius}) falls far to the right of this null histogram, the covariance difference is real; if it falls inside or near the histogram, the difference is no greater than what random grouping would produce.

\paragraph{Academic pairs.}
For CS vs.\ Shakespeare, $\tilde{F} = 1.238$ (GPT, $p{<}0.001$) and $1.267$ (BERT, $p{<}0.001$): the observed covariance difference lies far beyond the null distribution in all conditions (Figure~\ref{fig:frobenius}, top). For InfoSec vs.\ ToC, $\tilde{F} = 1.268$ (GPT) and $1.195$ (BERT), both $p{<}0.001$. The symmetric Gaussian KL decomposition reveals an informative structural shift: as the pair moves from Experiment~1 to Experiment~2, the covariance term accounts for a growing fraction of the total divergence (43\% $\to$ 44\% in GPT; 40\% $\to$ 47\% in BERT). As centroids converge, whatever distinction the model maintains is increasingly carried by the \emph{shape} of the activation distribution rather than its location.
\paragraph{Hate vs.\ no-hate speech.}
For GPT-OSS-20B, $\tilde{F} = 0.416$ ($p{=}0.025$)--marginal, with the red line sitting at the very right edge of the null histogram. For BERT-large-uncased, $\tilde{F} = 0.492$ ($p{=}0.130$)--not significant, with the red line inside the null distribution (Figure~\ref{fig:frobenius}, bottom). The covariance difference is $3{\times}$ smaller in magnitude than for the academic pairs, and critically, it is not robustly cross-model. The Gaussian KL decomposition for this pair is dominated by the covariance term (93--94\%), confirming that the first moment contributes essentially nothing.
\paragraph{Prompting as intervention.}
A natural question is whether the absence of separation is a fixed property of the representations or is sensitive to framing. We prepend a pair-specific classification prompt to each sentence (e.g., ``Classify this text as hate speech or not hate speech:'') and exclude prompt tokens from the mean pool, so the embedding reflects the sentence content processed through a model whose context has been set by the classification frame.
Prompting does not recover separation. For the hate pair, GPT's $\tilde{F}$ drops from 0.416 to 0.389 ($p{=}0.058$); BERT's from 0.492 to 0.483 ($p{=}0.165$). The mechanism is a homogenising attractor: transformer attention is context-sensitive, and prepending an identical prompt to all 200 sentences pulls all embeddings toward a shared ``classification-task'' region, compressing inter-class variance. For CS vs.\ Shakespeare, enough signal survives this compression ($\tilde{F}$ drops from 1.238 to 1.157, still $p{<}0.001$); for hate speech the prompt-induced compression erases what little remained.

The second-moment analysis confirms that the representational collapse observed for hate speech is not limited to class centroids--it extends to the full shape of the activation distributions. For well-separated domains, covariance structure differs significantly across classes. For hate speech, neither the location nor the shape of the distributions carries a reliable class-level signal. 

\section{Discussion}

This work examines whether neural networks organise their internal representations in ways that reflect conceptual separation. Across both CNNs and LLMs, representational separability broadly tracks how distinct concepts are to humans, and weakens as those distinctions become less clear.

For CNNs, familiar ImageNet classes form coherent and well-separated representations, whereas out-of-distribution concepts such as Rangoli and microscopy show substantially lower intra-class coherence. We also observe sensitivity to within-class domain shift: geographically different road images produce a measurable change in the activation blueprint, while cat-pose variation does not. Together, these results suggest that representation-space diagnostics can reveal both when an unseen concept is encoded inconsistently and when a familiar class shifts under a new domain.

For LLMs, a similar progression emerges. Clearly distinct domains remain well separated, related technical subdomains move closer together, and hate versus no-hate speech becomes nearly indistinguishable at the first moment. Second-moment analysis does not recover a reliable boundary across both models, and prompting fails to restore the lost structure. This suggests that, for ambiguous tasks, a downstream classifier may need to learn a decision boundary on top of representations that do not already encode a clear conceptual distinction.

The practical value of conceptual separation is therefore diagnostic. Before deploying a model on a new domain or an ambiguous task, one can ask whether target concepts are represented coherently and distinctly in its internal space. A lack of separation does not imply that classification is impossible, but it is evidence that the underlying representation deserves further investigation. We therefore view activation-space analysis as a useful complement to output-level evaluation, particularly for shifted, unfamiliar, or context-dependent concepts.

\section{Limitations and Future Work}
Our findings should be interpreted as indicative rather than conclusive. The empirical scope is limited to 100 samples per class and four representative architectures, and the hate-speech analysis relies on a single dataset from one platform and time, so its generalizability across taxonomies and linguistic contexts remains uncertain. Our analysis is limited primarily to the first two moments of the activation distributions and, for LLMs, only the final transformer layer, potentially missing structure present at intermediate layers. Our prompting intervention is also limited to a single classification-style template per concept pair; richer prompting strategies may produce different insights.

Several directions follow naturally from this work. First, these measures could be developed into practical \textit{diagnostic checks} for conceptual separation, assessing whether a model preserves coherent representations under new domains or ambiguous inputs before deployment. Developing a unified diagnostic metric will require broader validation across architectures, datasets, and tasks. A \textit{layer-wise} extension across transformer layers could reveal where conceptual separation emerges or collapses. We also plan to study \textit{adversarial perturbations}, testing whether representational coherence degrades further under targeted attack. More broadly, \textit{metric discovery} remains an open problem: identifying which geometric spaces best capture relational structure between concepts. Ultimately, this could enable \textit{metric imposition}, where training objectives or post-hoc regularisation encourage models to preserve desired conceptual relationships.

\bibliographystyle{unsrt}
\bibliography{references}

\newpage
\appendix

\section{Dataset Examples}
\label{app:dataset_examples}

To provide visual context for the datasets used in the CNN experiments, we show representative examples from each concept and domain considered in the study.

\subsection{In-Distribution Concepts}

\begin{figure}[H]
    \centering
    \includegraphics[width=\columnwidth]{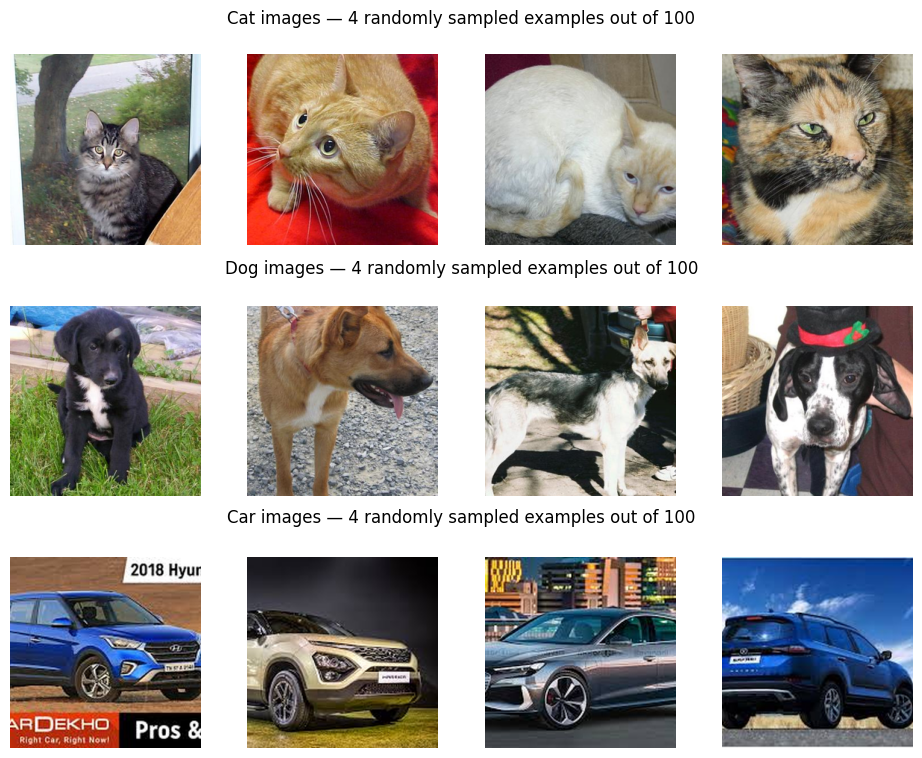}
    \caption{
    Representative cat, dog and car images used in the in-distribution experiments.
    }
    \label{fig:catdog_examples}
\end{figure}

Figure~\ref{fig:catdog_examples} shows representative cat, dog and car images used in the ImageNet-class experiments.

\subsection{OOD Concepts}

\begin{figure}[H]

    \centering
    \includegraphics[width=\columnwidth]{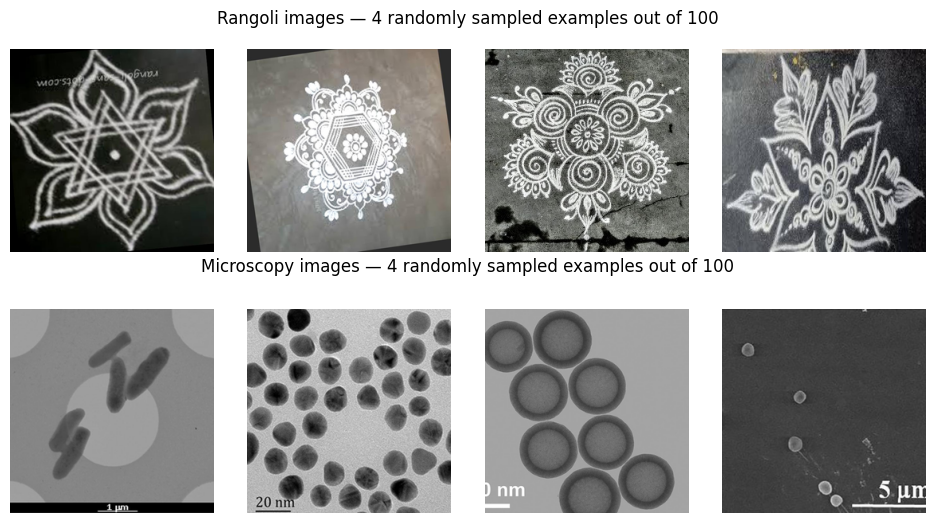}
    \caption{
    Representative Rangoli and microscopy images used in the unfamiliar-concept experiments.
    }
    \label{fig:ood_examples}
\end{figure}

Figure~\ref{fig:ood_examples} shows representative Rangoli and microscopy images used in the unfamiliar-concept experiments. Although both form visually coherent human-defined categories, neither corresponds to an explicit ImageNet class.

\subsection{Within-Class Distribution Shift}

\begin{figure}[H]
    \centering
    \includegraphics[width=\columnwidth]{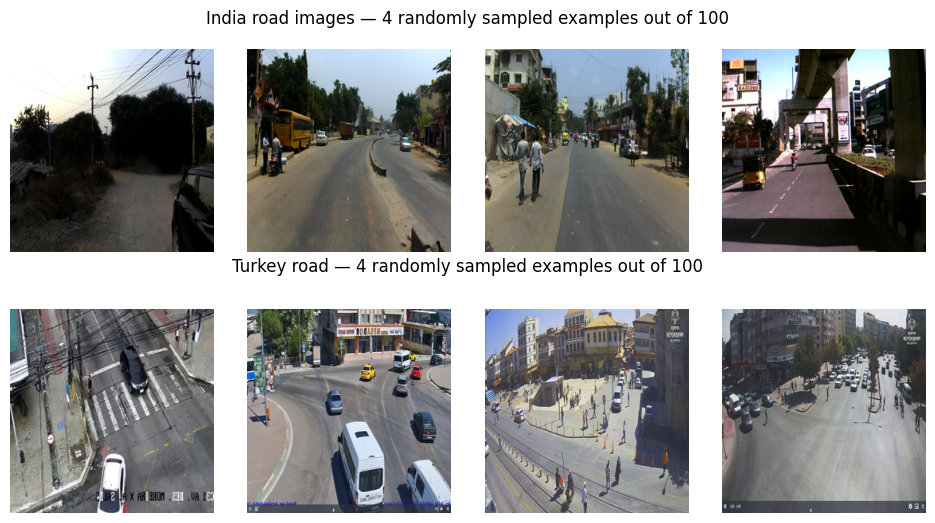}
    \caption{
    Representative road images from India and Turkey used in the within-class distribution-shift experiment.
    }
    \label{fig:road_examples}
\end{figure}

Figure~\ref{fig:road_examples} shows representative road images from India and Turkey held out test sets used to study within-class geographic distribution shift.

\subsection{Within-Class Pose Variation}

\begin{figure}[H]
    \centering
    \includegraphics[width=\columnwidth]{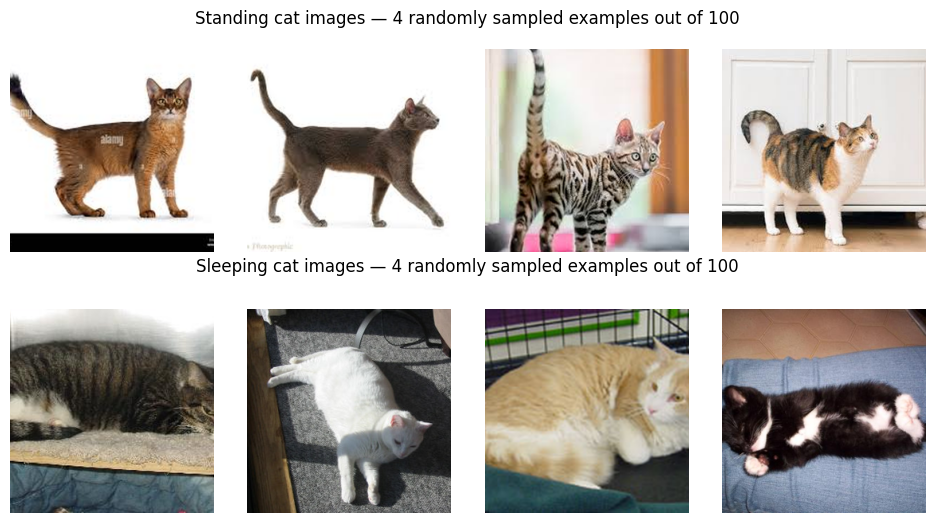}
    \caption{
    Representative sleeping and standing cat images used in the pose-variation experiment.
    }
    \label{fig:pose_examples}
\end{figure}

Figure~\ref{fig:pose_examples} shows representative sleeping and standing cat images from the held out test sets. These images were manually picked from the cat dataset \cite{jacob_catdog}.

\section{PCA and Distance-Based Analysis}
\label{app:cnn_distances}

\begin{figure}[H]
\centering
\includegraphics[width=\columnwidth]{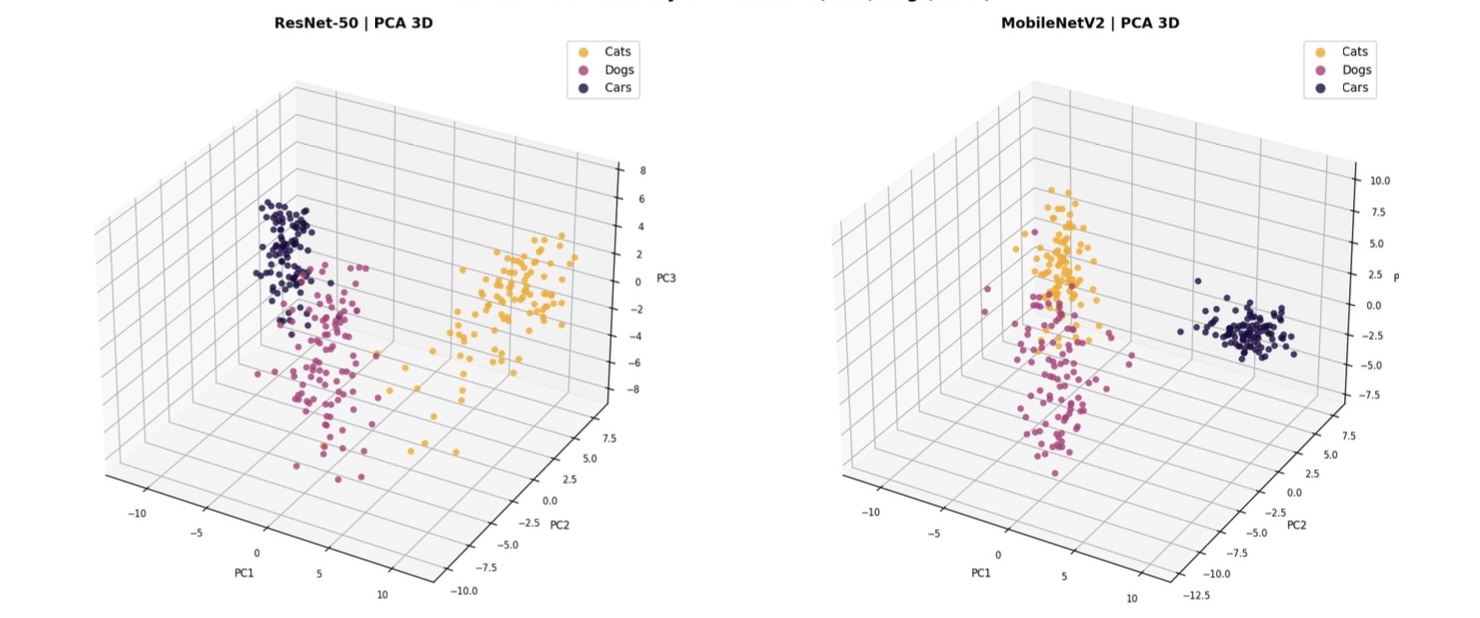}
\caption{PCA projections of cat, dog, and car representations for ResNet-50 and MobileNetV2.}
\label{fig:appendix_pca}
\end{figure}

\begin{figure*}[t]
\centering
\includegraphics[width=0.92\columnwidth]{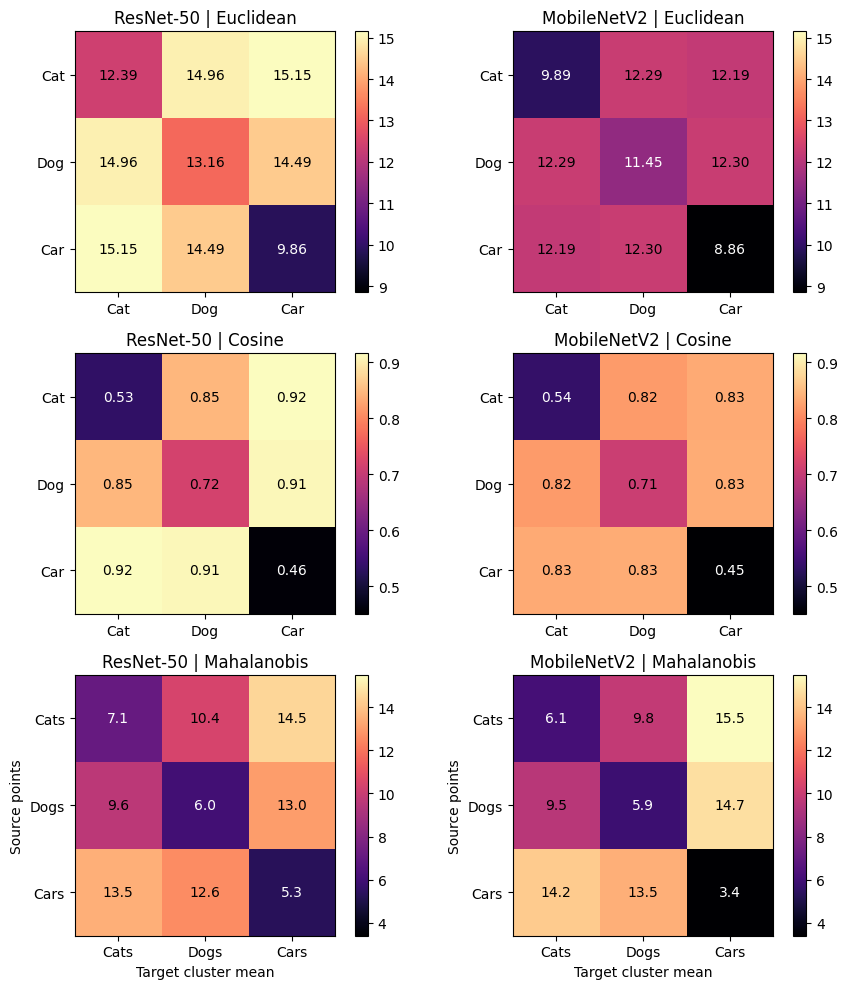}
\caption{Euclidean, cosine and Mahalanobis distances between cat, dog, and car representations. Intra-class distances are consistently smaller than inter-class distances.}
\label{fig:appendix_distances}
\end{figure*}

Across both architectures, cats, dogs, and cars form distinct regions in PCA space (Figure~\ref{fig:appendix_pca}). Cosine and Mahalanobis distances are also smaller within classes than across classes, with cats and dogs closer to each other than either is to cars (Figure~\ref{fig:appendix_distances}). This is consistent with the KL analysis in the main paper and indicates that the expected semantic structure is present across several measures of representation geometry.

\section{Prediction Consistency for Out-of-Distribution Concepts}
\label{app:ood_predictions}

We further examine the models' ImageNet predictions as a behavioural view of representational consistency. The goal is not to expect correct predictions for rangoli or microscopy, since neither is an ImageNet class. Instead, we ask whether different images from the same concept are mapped to similar known classes.

\begin{figure*}[t]
\centering
\includegraphics[width=\columnwidth]{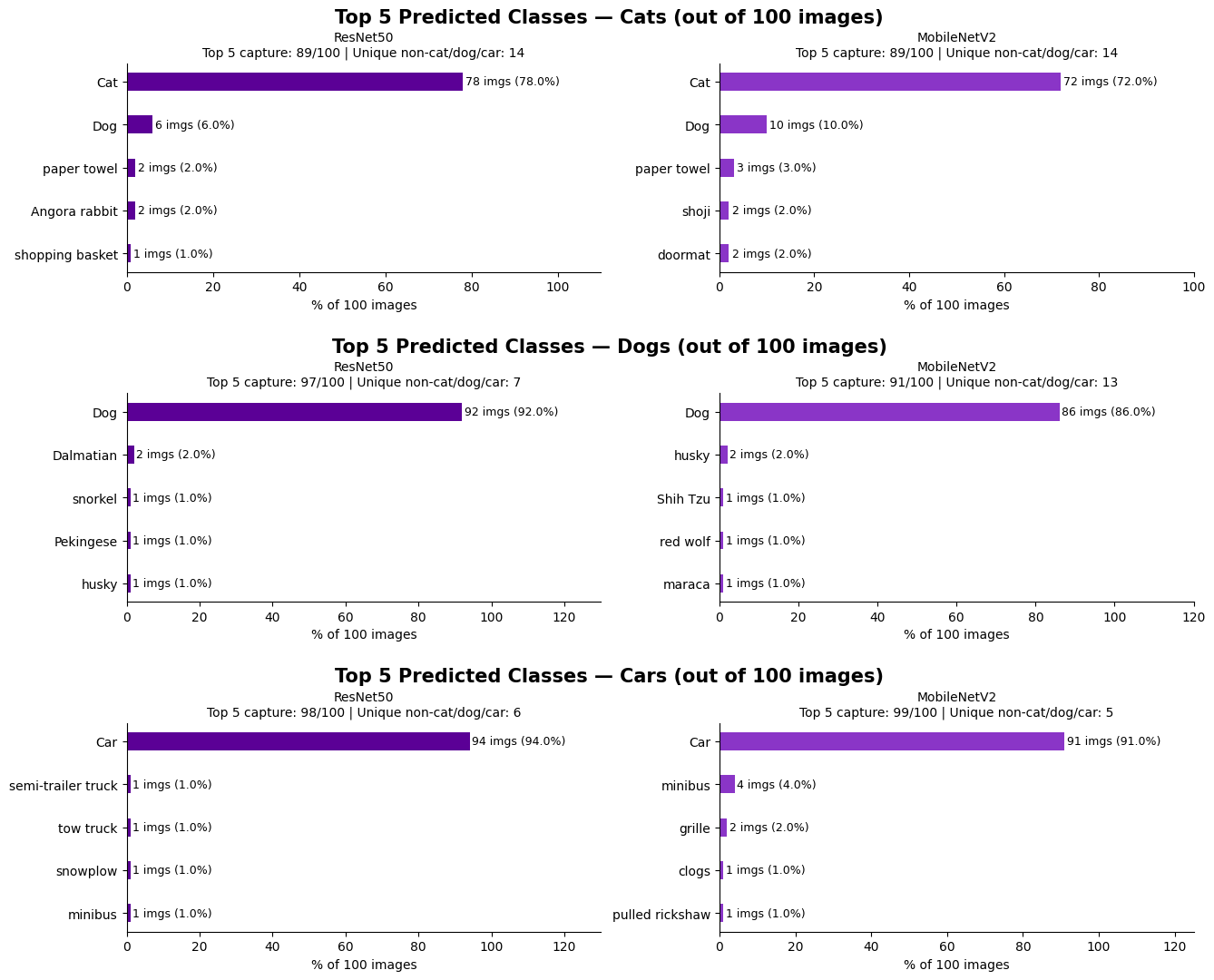}
\caption{Top ImageNet predictions for cats, dogs, and cars. Predictions show relatively low variation within each concept.}
\label{fig:appendix_id_predictions}
\end{figure*}

For cats, dogs, and cars, the predictions are strongly concentrated in the expected semantic category. For example, over $75\%$ of cat images are predicted as cats, while the corresponding numbers rise to around 90\% for dogs and cars. Even when the exact ImageNet label differs, most remaining predictions are closely related classes. For readability, we therefore collapse fine-grained labels such as individual cat breeds, dog breeds, and car types into broader semantic groups. The resulting prediction distributions are highly concentrated (Figure~\ref{fig:appendix_id_predictions}), which is consistent with the low intra-class KL divergence observed for these concepts.

\begin{figure}[H]
\centering
\includegraphics[width=\columnwidth]{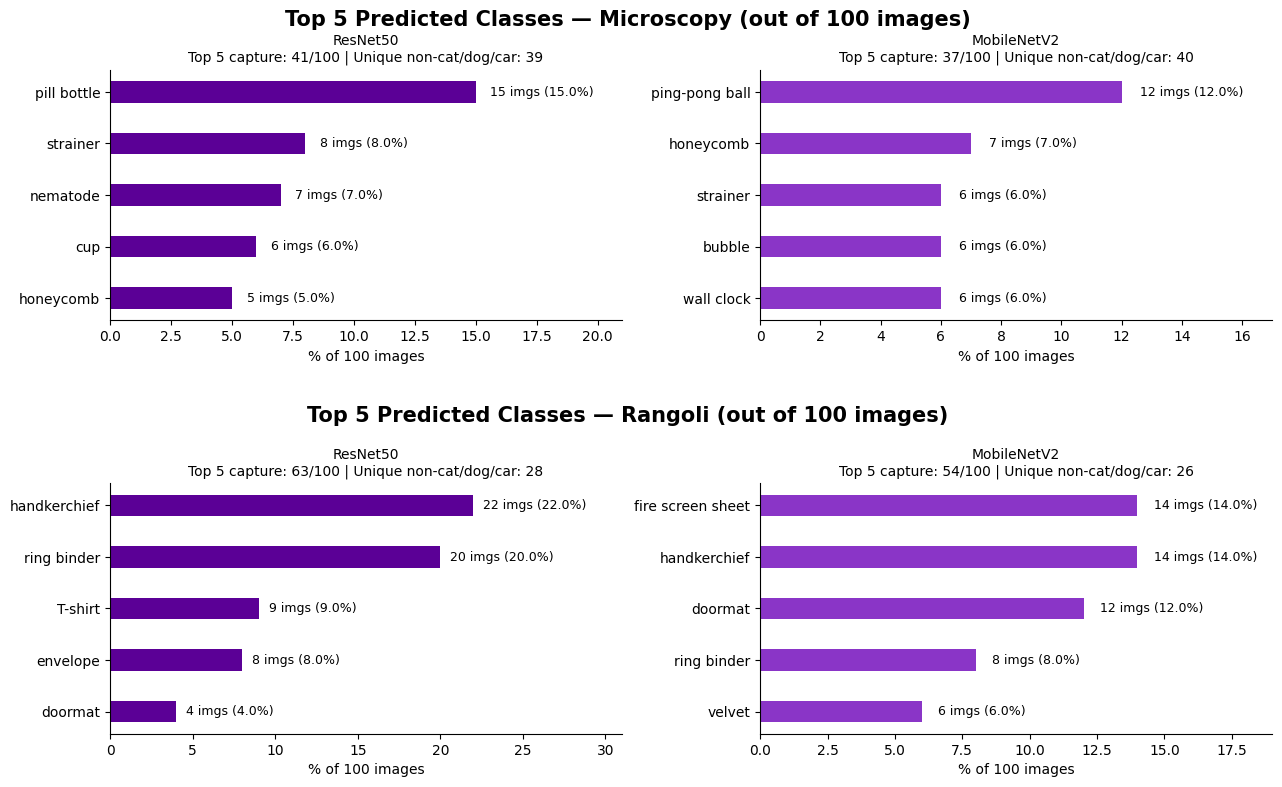}
\caption{Top ImageNet predictions for rangoli and microscopy images. Images from the same concept are mapped across a diverse set of ImageNet classes.}
\label{fig:appendix_ood_predictions}
\end{figure}

Rangoli and microscopy show the opposite pattern. For microscopy, the most common prediction accounts for only 15\% of ResNet-50 images and 12\% of MobileNetV2 images; for Rangoli, the corresponding values are 22\% and 14\%. The top five predicted classes together capture only 37--63\% of the images, compared with 89--99\% for cats, dogs, and cars. Predictions are therefore spread across many unrelated ImageNet classes rather than concentrated around a single semantic group (Figure~\ref{fig:appendix_ood_predictions}). This is not a classification error in the usual sense, since neither Rangoli nor microscopy has a correct ImageNet label. Instead, the sharp drop in prediction consistency provides complementary evidence that different examples from the same human-defined concept elicit less coherent model responses, in line with the higher intra-class KL divergence reported in the main paper.

\section{Representative examples from the LLM datasets}
\label{app:llm-examples}

We list three sentences from each concept class used in the language-model experiments.
The full text collections are included in the public code repository under \texttt{data/llm\_data/}.
The academic concept pairs (Computer Science vs.\ Shakespeare and Information Security vs.\ Theory of Computation) consist of short definitional or canonical sentences compiled from Wikipedia and other publicly available sources.
Hate and no-hate examples are social-media posts drawn from the 2020 U.S.\ election Twitter corpus of Grimminger and Klinger~\cite{grimminger2021hate}, included in the public code repository under \texttt{data/llm\_data/}, and are reproduced as in the source (including hashtags and usernames). They may contain profanity and partisan language.

\begin{table}[H]
\centering
\caption{Three examples from each LLM concept class}
\label{tab:app-llm-examples}
\small
\begin{tabularx}{\textwidth}{@{}lX@{}}
\toprule
\textbf{Class} & \textbf{Examples} \\
\midrule
Shakespeare
& \begin{enumerate}[nosep,leftmargin=1.4em]
    \item To be, or not to be: that is the question.
    \item Brevity is the soul of wit.
    \item All the world's a stage, and all the men and women merely players.
  \end{enumerate} \\
\midrule
Computer Science
& \begin{enumerate}[nosep,leftmargin=1.4em]
    \item An algorithm is a finite sequence of well-defined instructions to solve a problem.
    \item Time complexity measures how an algorithm's runtime grows with input size.
    \item Big-O notation describes the upper bound of an algorithm's growth rate.
  \end{enumerate} \\
\midrule
Information Security
& \begin{enumerate}[nosep,leftmargin=1.4em]
    \item Information security ensures confidentiality, integrity, and availability of data.
    \item Encryption transforms plaintext into unreadable ciphertext.
    \item Asymmetric encryption uses a public-private key pair.
  \end{enumerate} \\
\midrule
Theory of Computation
& \begin{enumerate}[nosep,leftmargin=1.4em]
    \item Theory of Computation studies abstract models of computation.
    \item Deterministic finite automata (DFA) have exactly one transition per input symbol.
    \item Regular languages are recognized by finite automata.
  \end{enumerate} \\
\midrule
Hate Speech
& \begin{enumerate}[nosep,leftmargin=1.4em]
    \item Worship of dictator trump.
    \item Give me 10 GOOD fucking reasons to vote trump. \#BIDEN2020
    \item Don't let your daddy issues influence you to vote for Biden \#magaszn \#maga2020
  \end{enumerate} \\
\midrule
No-Hate Speech
& \begin{enumerate}[nosep,leftmargin=1.4em]
    \item Yay! Voted today! \#BidenHarris2020
    \item Voted today! Please find the time and vote. \#vote \#BidenHarris2020
    \item Got my \#Trump2020 swag in today!!!!!! So excited!
  \end{enumerate} \\
\bottomrule
\end{tabularx}
\end{table}

\section{LLM Distance Matrices and PCA Projections}
\label{app:llm_distance}

This appendix provides the full set of distance matrices, six-category KL matrices, Frobenius null distributions, and moment decomposition analyses across both GPT-OSS-20B and BERT-large-uncased. 

\begin{figure}[H]
    \centering
    \includegraphics[width=0.75\linewidth]{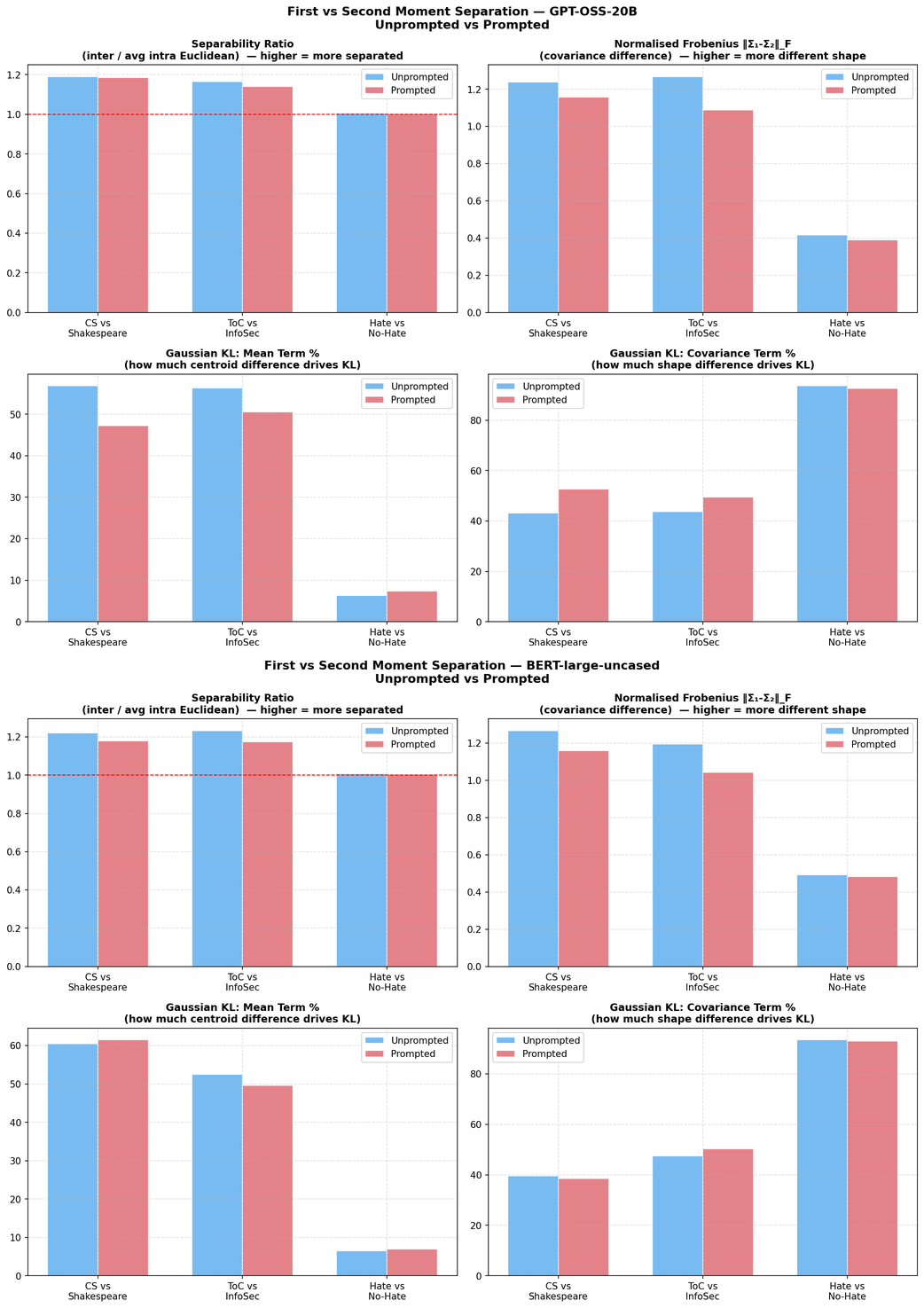}
    \caption{First vs.\ second moment separation and Gaussian KL decomposition under unprompted and prompted conditions. Top: GPT-OSS-20B; Bottom: BERT-large-uncased. Separability ratios collapse to unity and Gaussian KL is overwhelmingly dominated by the covariance term ($>90\%$) for the hate speech distinction across both architectures.}
    \label{fig:app_moment_summary}
\end{figure}

\begin{figure}[htb!]
    \centering
    \includegraphics[width=0.9\linewidth]{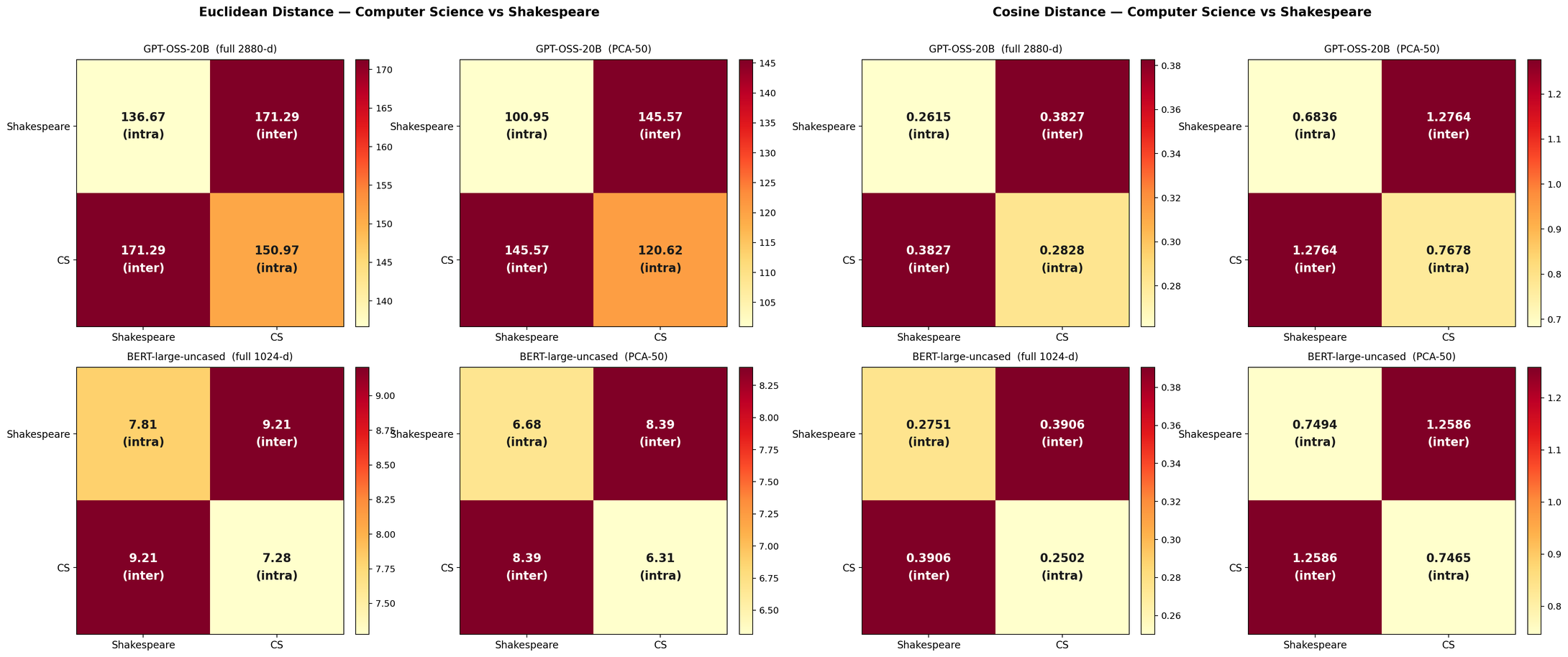}
    \caption{Distance matrices for CS vs.\ Shakespeare. Left: Euclidean; Right: Cosine. Within each block: GPT (top) and BERT (bottom).}
    \label{fig:app_dist_cs}
    
    \vspace{1.5em}
    \includegraphics[width=0.9\linewidth]{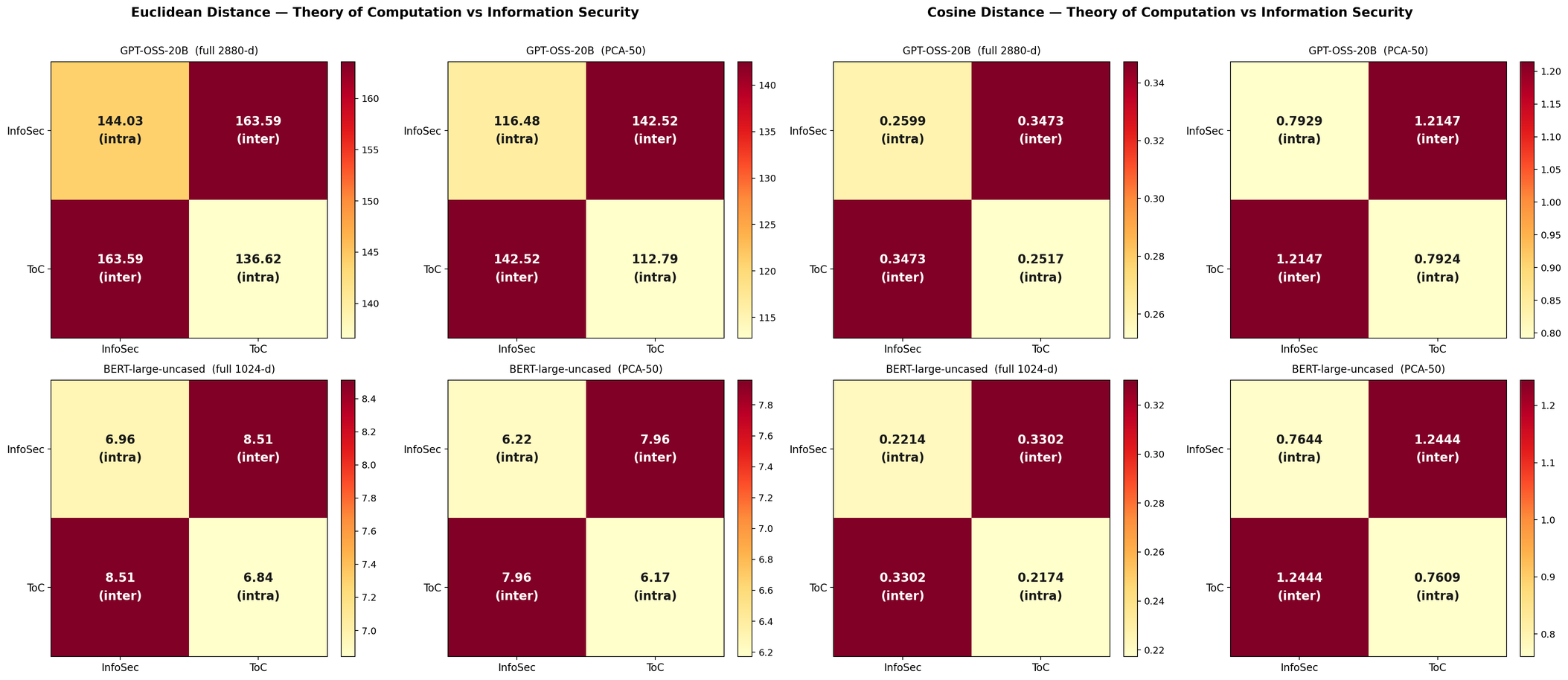}
    \caption{Distance matrices for InfoSec vs.\ Theory of Computation. Layout as in Figure~\ref{fig:app_dist_cs}.}
    \label{fig:app_dist_infosec}
    
    \vspace{1.5em}
    \includegraphics[width=0.9\linewidth]{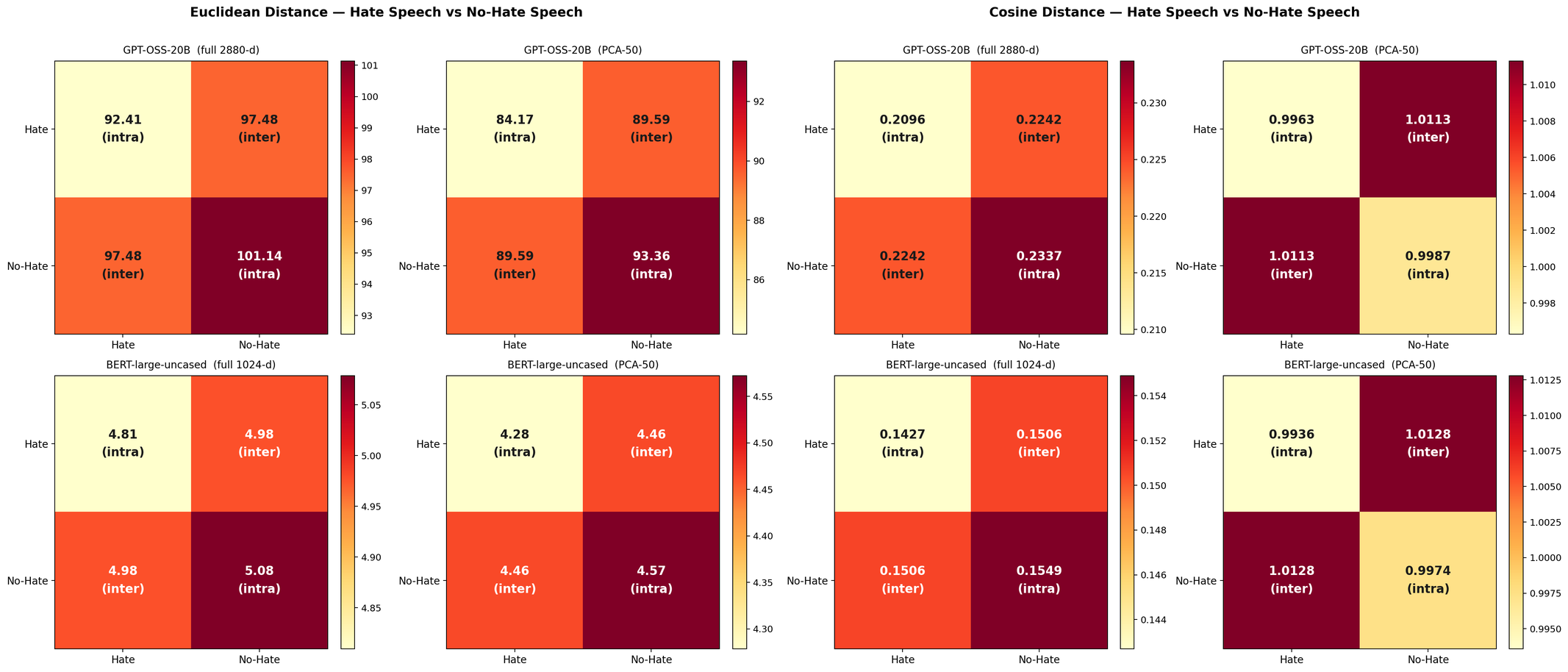}
    \caption{Distance matrices for Hate Speech vs.\ No-Hate Speech. Layout as in Figure~\ref{fig:app_dist_cs}.}
    \label{fig:app_dist_hate}
\end{figure}

\begin{figure}[htb!]
    \centering
    \includegraphics[width=0.95\linewidth]{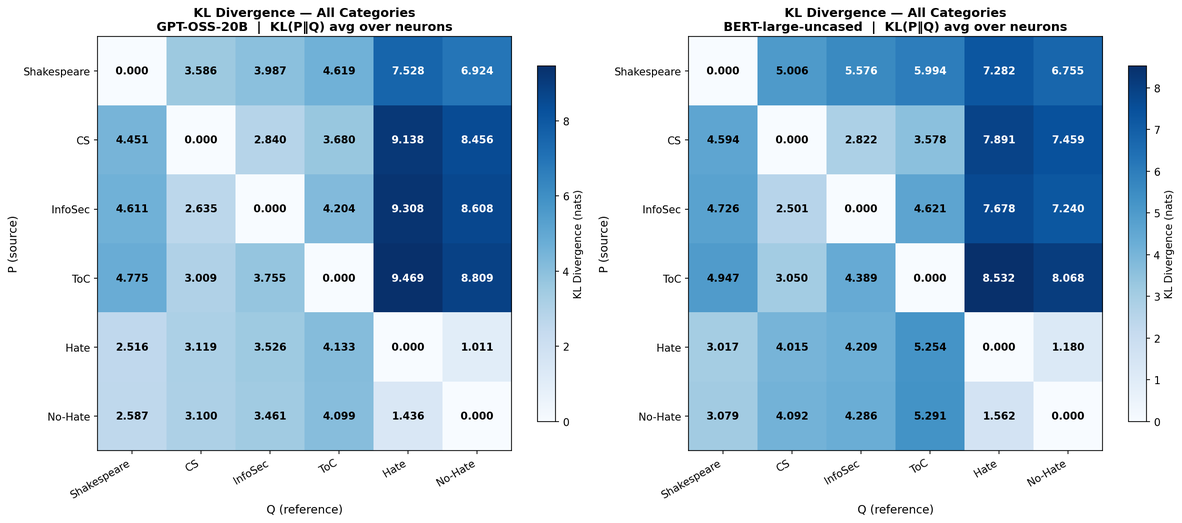}
    \caption{Per-neuron KL divergence matrix across all six categories (Left: GPT; Right: BERT). Hate and No-Hate Speech have the smallest off-diagonal KL of any pair.}
    \label{fig:app_kl_6cat}

    \vspace{1em}
    \includegraphics[width=0.8\linewidth]{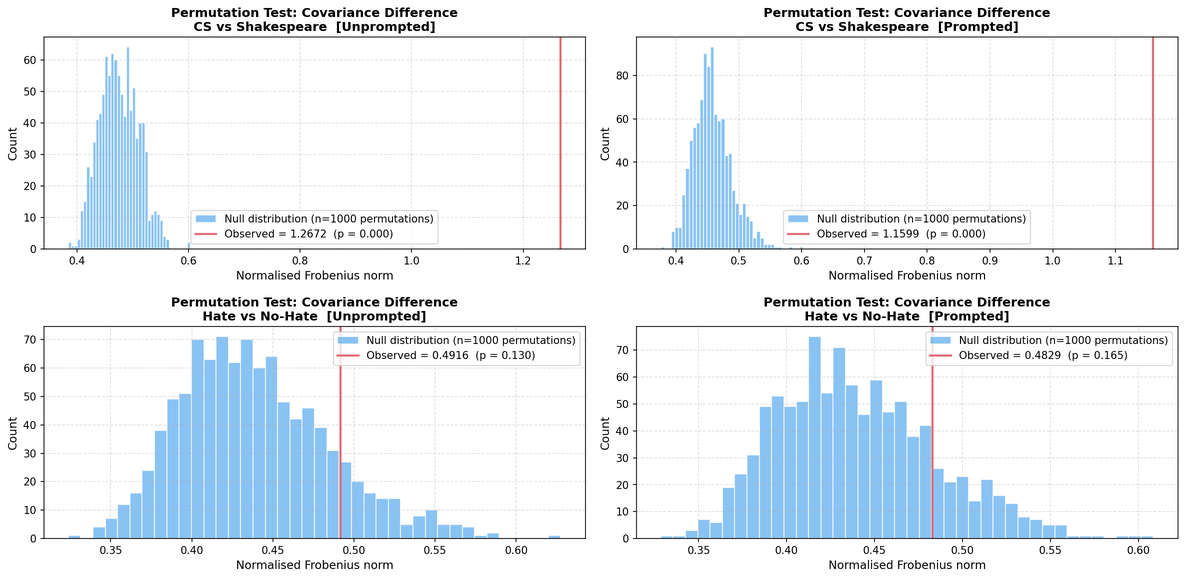}
    \caption{Permutation test null distributions (BERT-large-uncased) of the normalised Frobenius norm $\tilde{F}$ for CS vs.\ Shakespeare (top) and Hate Speech (bottom).}
    \label{fig:app_frob_bert}
    
    \vspace{1em}
    \includegraphics[width=0.8\linewidth]{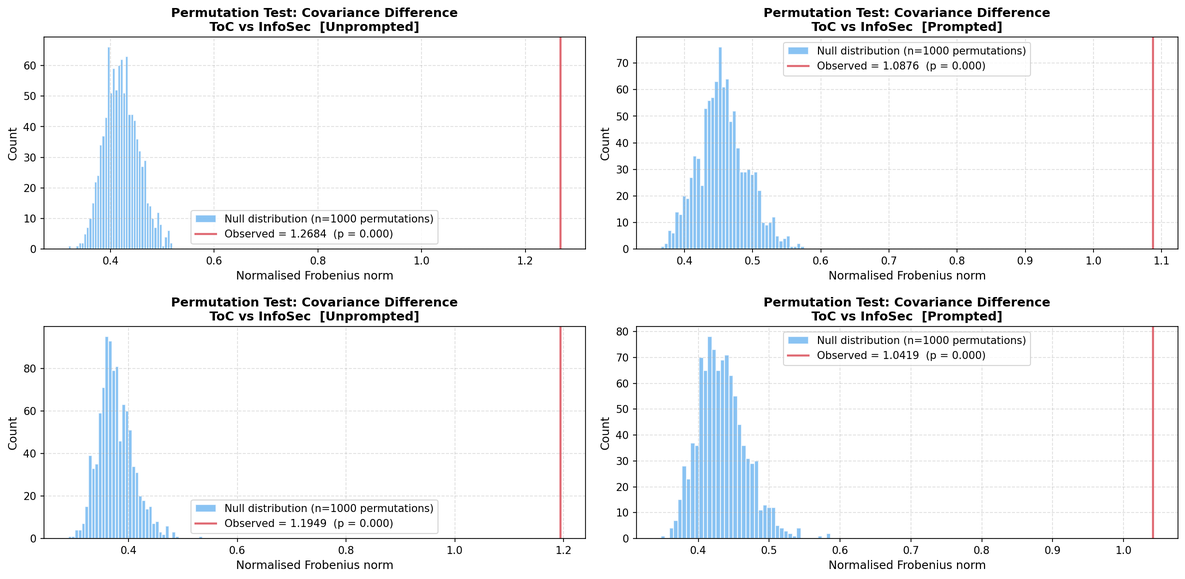}
    \caption{Permutation test null distributions for InfoSec vs.\ Theory of Computation. Top: GPT-OSS-20B; Bottom: BERT-large-uncased. Left: Unprompted; Right: Prompted.}
    \label{fig:app_frob_infosec}
\end{figure}

\section{Hate Speech Data and Labelling Framework}
\label{app:hate_data}

For hate speech, 100 sentences per class are drawn from the 2020 U.S.
presidential-election Twitter corpus of Grimminger and Klinger
\cite{grimminger2021hate}. We derive the binary Hate Speech / No-Hate Speech
partition from the corpus's hate/offensive-speech annotations.

The choice of a single, domain-controlled corpus is deliberate. Cross-domain hate speech detection is a fundamentally different problem from within-domain detection, and a dataset drawn from multiple platforms, time periods, or political contexts would conflate domain shift with the conceptual distinction we aim to isolate. By restricting to tweets from the same event window and platform, any observed difference in embedding geometry reflects a difference in the underlying concept--the linguistic intent and framing of the speech--rather than writing style, topic distribution, or platform conventions.

This design connects to the broader challenge of hate speech annotation. Unlike academic domain labels, which can be assigned with high inter-annotator agreement, hate speech labels are socially contested: annotators disagree on boundary cases depending on cultural context, personal sensitivity, and the operational definition used~[1, 2]. The geometric analysis in the main paper sidesteps the labelling problem by asking whether pre-trained representations encode \emph{any} concept-level distinction between these classes, regardless of where the decision boundary is drawn. The finding that they do not--at either the first or second moment--suggests that the difficulty of hate speech annotation may have a structural counterpart in the representational geometry of general-purpose language models.

\end{document}